\documentclass[letterpaper]{article} 
\usepackage{aaai25}  
\usepackage{times}  
\usepackage{helvet}  
\usepackage{courier}  
\usepackage[hyphens]{url}  
\usepackage{graphicx} 
\usepackage{natbib}  
\usepackage{caption} 
\usepackage{algorithm}
\usepackage{algorithmic}
\usepackage{amsfonts}       
\usepackage{nicefrac}       
\usepackage{microtype}      
\usepackage{amsmath}        
\usepackage{booktabs}       
\usepackage{multirow}
\usepackage{caption}
\usepackage{subcaption}
\usepackage{enumitem}
\usepackage[table]{xcolor}
\usepackage{colortbl} 
\usepackage{array}

\usepackage{newfloat}
\usepackage{listings}
\DeclareCaptionStyle{ruled}{labelfont=normalfont,labelsep=colon,strut=off} 
\floatstyle{ruled}
\newfloat{listing}{tb}{lst}{}
\floatname{listing}{Listing}
\title{\textcolor{red}{ConDo}: \textcolor{red}{Con}tinual \textcolor{red}{Do}main Expansion for Absolute Pose Regression}
\author{
    Zijun Li\textsuperscript{\rm 1}\equalcontrib,
    Zhipeng Cai\textsuperscript{\rm 2}\equalcontrib\thanks{Corresponding authors.},
    Bochun Yang\textsuperscript{\rm 1},
    Xuelun Shen\textsuperscript{\rm 1}, \\
    Siqi Shen\textsuperscript{\rm 1},
    Xiaoliang Fan\textsuperscript{\rm 1},
    Michael Paulitsch\textsuperscript{\rm 2},
    Cheng Wang\textsuperscript{\rm 1}\footnotemark[2]
}

\affiliations {
    \textsuperscript{\rm 1}Fujian Key Laboratory of Sensing and Computing for Smart Cities, Xiamen University, China\\
    \textsuperscript{\rm 2}Intel Labs \\
    \{lizijun;yangbc;xuelun\}@stu.xmu.edu.cn,\{zhipeng.cai;michael.paulitsch\}@intel.com,\{siqishen;fanxiaoliang;cwang\}@xmu.edu.cn
}

\begin{document}

\maketitle

\begin{abstract}
Visual localization is a fundamental machine learning problem. Absolute Pose Regression (APR) trains a scene-dependent model to efficiently map an input image to the camera pose in a pre-defined scene. However, many applications have continually changing environments, where inference data at novel poses or scene conditions (weather, geometry) appear after deployment. Training APR on a fixed dataset leads to overfitting, making it fail catastrophically on challenging novel data. This work proposes Continual Domain Expansion (ConDo), which continually collects unlabeled inference data to update the deployed APR. Instead of applying standard unsupervised domain adaptation methods which are ineffective for APR, ConDo effectively learns from unlabeled data by distilling knowledge from scene-agnostic localization methods. By sampling data uniformly from historical and newly collected data, ConDo can effectively expand the generalization domain of APR. Large-scale benchmarks with various scene types are constructed to evaluate models under practical (long-term) data changes. ConDo consistently and significantly outperforms baselines across architectures, scene types, and data changes. On challenging scenes (Fig.~\ref{Fig-teaser}), it reduces the localization error by $>7$x ($14.8$m vs $1.7$m). Analysis shows the robustness of ConDo against compute budgets, replay buffer sizes and teacher prediction noise. Comparing to model re-training, ConDo achieves similar performance up to 25x faster. 
\end{abstract}
\begin{links}
    \link{Code}{https://github.com/ZijunLi7/ConDo}
\end{links}

\section{Introduction}\label{sec:intro}
Localizing an image in a given scene is a fundamental machine learning problem. The scene is defined by a set of \emph{reference} images with known camera poses, and the task is to return the camera pose of a query image.

Different types of methods have been developed for visual localization. Retrieval-based methods search for reference images similar to the input and use their poses as the output~\citep{torii201524,arandjelovic2016netvlad}. These methods require storing the reference images during inference, which introduces memory overheads. There are also methods applying explicit geometric optimizations to obtain more fine-grained poses~\citep{sarlin2019coarse,hyeon2021pose,kim2023ep2p}. Though more accurate, geometric optimization introduces computational overheads, making them limited when facing real-time applications.
\begin{figure}[!t]
    \centering
    \includegraphics[width=0.9\linewidth]{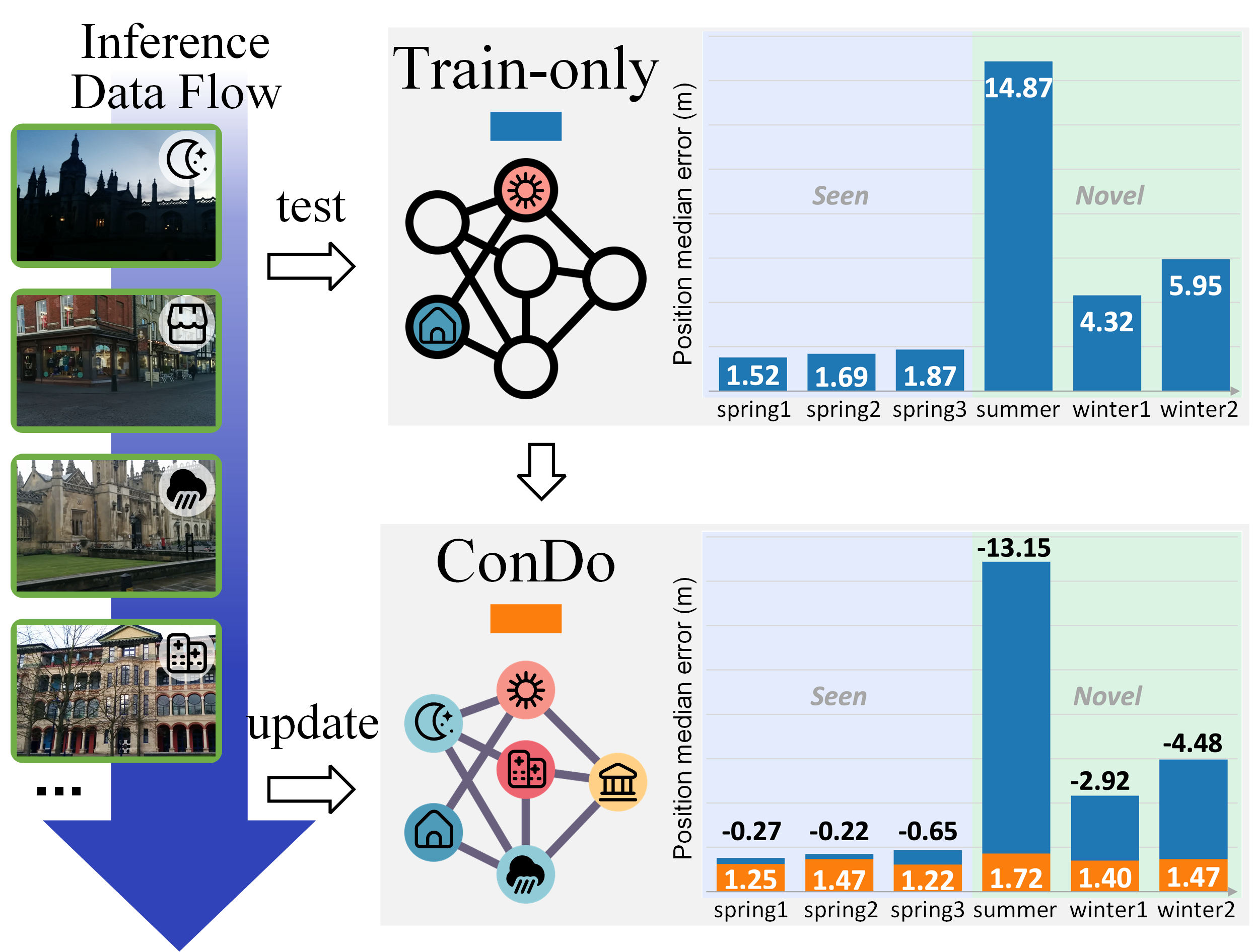}
    \caption{Teaser. We propose Continual Domain Expansion (ConDo) for APR, which utilizes unlabeled data seen during inference to expand the generalization domain of APR. Novel benchmarks are proposed to study practical scenarios where images are captured at novel poses or continually changing environments (left). The x-axis of histograms represents test data from various scans and y-axis indicates the estimated position median error. Trained only on data from spring, the deployed APR cannot handle summer and winter data (top). ConDo updates the model continually with unlabeled inference data and limited computation budgets, effectively expanding the generalization domain over time  (bottom).}
    \label{Fig-teaser}
\end{figure}
Absolute pose regression (APR)~\citep{kendall2015posenet,brahmbhatt2018geometry} is an important type of visual localization methods. It trains a light-weight scene-dependent neural network to directly output the camera pose of the query image. Such direct image-to-pose mapping makes APR highly efficient for both computation and memory (see Appendix~\ref{sec:memory-time} for comparisons), suitable for real-time applications on edge devices. Comparing to multi-view methods like SLAM~\cite{campos2021orb}, APR can derive camera poses from a single image. Though with clear advantages, the scene-dependent model training also limits the robustness of APR on novel data seen during inference~\citep{sattler2019understanding}. The novel data can be captured either at poses distant from the training data~\citep{sattler2019understanding}, or with unseen lighting, weather, or geometry (new construction) conditions caused by the change of time~\citep{cai2023clnerf}. Fig.~\ref{Fig-teaser} shows an example (top) where the model only trained on data captured in spring sees inference data from summer and winter. The accuracy dropped heavily even though different trajectories have similar pose distributions.

A naive solution for this problem is to obtain new data with ground truth (GT) that cover novel poses and scene conditions, train a new model from scratch on both historical and new data, and then deploy the new model for inference. However, obtaining ground-truth data for APR often requires manual scene traverses with 3D scanners, which not only introduces extra laboring costs but also cannot guarantee to cover all novel data in a continually changing environment. Meanwhile, re-training models with more data needs more computation and time to converge. 

In this work, we propose \emph{\textbf{Con}tinual \textbf{Do}main Expansion (ConDo)} for APR. ConDo leverages unlabeled data seen after model deployment to continually and efficiently update APR. Though unsupervised domain adaptation methods have been proposed for standard classification/regression tasks, as shown later in the experiments, they struggle to generate effective supervision signals for APR. Inspired by the fact that scene-agnostic methods~\citep{arandjelovic2016netvlad,sarlin2019coarse,von2022dm,campos2021orb} are much more robust to scene and pose changes, we instead generate supervision signals on unlabeled data by distilling knowledge from them. As shown in Fig.\ref{Fig-teaser}, this simple yet effective strategy improves not only the performance on data from the same domain, but also the general robustness of APR, leading to better performance on other domains. Meanwhile, the model is updated continually without re-training, so that the computation does not grow over time. Besides the case of a single scene, the multi-head architecture is applied to make ConDo also applicable to sequentially revealed new scenes with a minimal model parameter increase. To thoroughly evaluate APR on data with practical changes, we construct benchmarks that cover 1) indoor and outdoor scenes, 2) large-scale city-level data, 3) (long-term) scene changes and novel camera poses.

Experiments validate the effectiveness of ConDo on different baseline architectures and data with both scene and pose changes. It reduces the localization error by more than an order of mangnitude on challenging data. Comprehensive analysis shows the robustness of ConDo w.r.t. the knowledge distillation teacher, replay buffer sizes, compute budgets and so on. Comparing to model re-training, ConDo can reach similar performance with up to 25x compute/time reduction.

\section{Related Work}\label{sec:related_work}

\noindent\textbf{Absolute pose regression.} APR is a classical visual localization approach, which directly regresses the camera's pose based on a single input image when revisiting a known environment. \citep{kendall2015posenet} proposed the first APR method, which contains a feature extractor and pose regressor in the architecture. Follow-up methods improve the performance by introducing attention layers~\citep{wang2020atloc}, Transformers~\citep{shavit2021learning} and Diffusion models~\citep{wang2023robustloc}. 
To better leverage scene information, visual odometry and motion constraints~\citep{brahmbhatt2018geometry,xue2019local} have been introduced. Recently, NeRFs (Neural Radiance Fields) have been used to generate more data~\citep{moreau2022lens,chen2022dfnet} or geometric constraints ~\citep{chen2021direct,Moreau_2023_ICCV} for APR training. Though efficient, APR struggles to generalize to novel poses~\citep{sattler2019understanding} and scene changes~\citep{cai2023clnerf}. ConDo is designed to address this problem.

\noindent\textbf{Continual learning and other related problems.} Conventional continual learning methods~\citep{kirkpatrick2017overcoming,aljundi2017expert} aim to prevent catastrophic forgetting with limited storage. Recent approaches~\citep{cai2021online,prabhu2023online} switch the focus to limited computation, aiming to achieve fast adaptation under practical limitations of training resources. This setup is similar to ConDo except that the ground truth labels are assumed to be available during continual model updates, which is impractical for localization systems that require high-end scanning devices to obtain accurate labels. Unsupervised domain adaptation (UDA)~\citep{chen2021representation,nejjar2023dare} aims to adapt a pre-trained model to a target domain without ground truth. Forgetting and computation budgets are not the major concern. Meta-learning~\citep{finn2017model} trains on diverse tasks to adapt with GT labels during inference, both of which are difficult to obtain for APRs. ConDo aims to continually adapt to new domains while preserving the performance of old ones. ConDo distills knowledge from scene-independent localization methods, which is more effective than standard UDA methods for APR.

\section{Method}\label{sec:method}
\begin{figure*}[!h]
    \centering
    \includegraphics[width=0.85\linewidth]{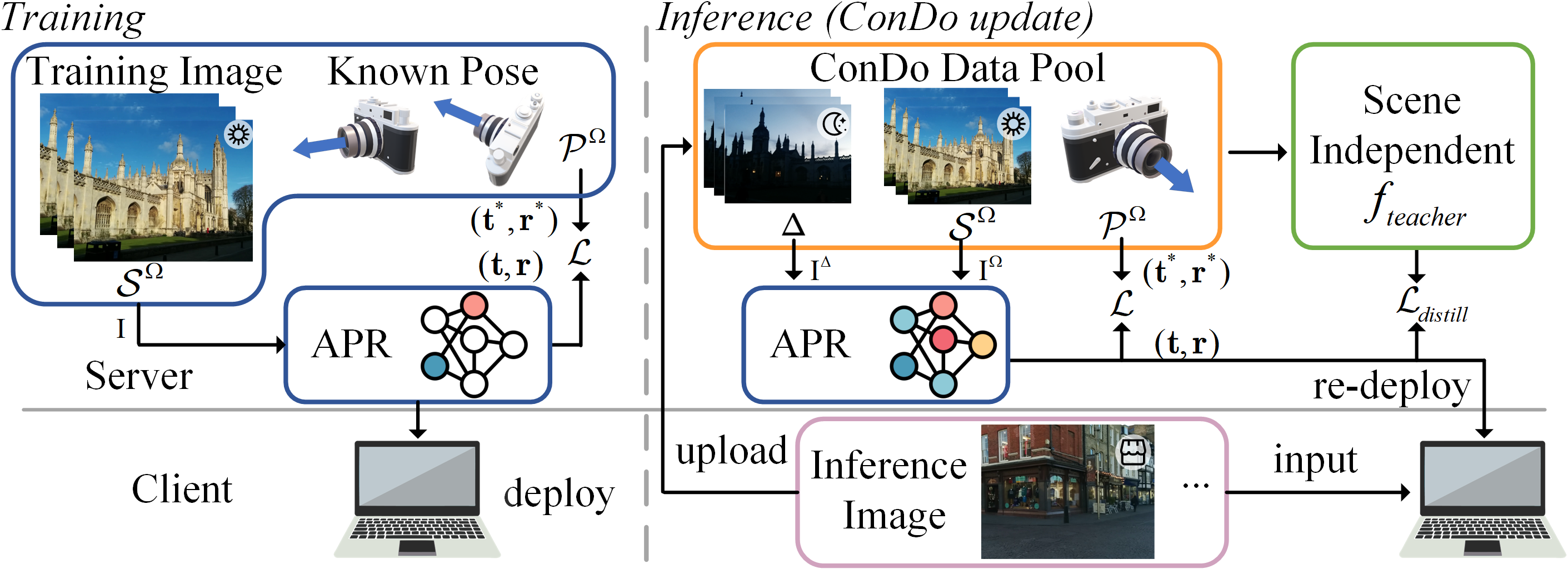}
    \caption{ConDo Pipeline. Left: After the normal APR training on labeled data, the model is deployed to the client. Right: After deployment, the client uploads the unlabeled data to the server. The server continually expands the generalization domain of APR by updating it with the labeled training data $(\mathcal{S}^\Omega, \mathcal{P}^\Omega)$, unlabeled data $\Delta$ and a scene-independent teacher method $f_\text{teacher}$ for knowledge distillation. Limited computation is assigned to each round of model update to ensure practical efficiency.}\label{Fig-pipeline}
\end{figure*}
\subsection{Preliminaries}\label{sec:prelim}
Given an image $\mathbf{I} \in \mathbb{R}^{H\times W\times C}$, APR~\citep{kendall2015posenet} learns a function $\mathbf{t}, \mathbf{r} = f(\mathbf{I} | \boldsymbol{\theta})$ parametrized by the neural network weights $\boldsymbol{\theta}$ that maps $\mathbf{I}$ to the camera position $\mathbf{t}\in\mathbb{R}^3$ and orientation $\mathbf{r}\in\mathbb{R}^4$ in a pre-defined scene $\Omega$. $\Omega$ is defined by a set of training images $\mathcal{S}^\Omega = \{\mathbf{I}^\Omega_i\}_{i=1}^N$ with known poses $\mathcal{P}^\Omega = \{\mathbf{t}^\Omega_i, \mathbf{r}^\Omega_i\}_{i=1}^N$. The function $f$ is a neural network commonly comprised of a feature extractor $g$ and a regressor $h$, i.e., $f=h\circ g$ where $g$ extracts the image level feature and $h$ projects the extracted feature to $\mathbf{t}$ and $\mathbf{r}$. Conventional APR frameworks train models on $\mathcal{S}^\Omega$ and $\mathcal{P}^\Omega$ with the regression loss~\citep{kendall2017geometric}:
\begin{equation}\label{eq:loss}
    \mathcal{L}(\mathbf{I}, \mathbf{t}^*, \mathbf{r}^*) = \| \mathbf{t} - \mathbf{t}^* \|e^{-s_t} + s_t + \| \mathbf{r} - \frac{\mathbf{r}^*}{\|\mathbf{r}^*\|} \|e^{-s_r} + s_r,
\end{equation}
where $\mathbf{t}$ and $\mathbf{r}$ are the predicted pose on $\mathbf{I}$, $(\mathbf{t}^*$, $\mathbf{r}^*) \in \mathcal{P}^\Omega$ are the ground truth, and $s_t$ and $s_r$ are learnable parameters to balance the position and orientation losses. After training, the APR model is deployed to the environment for inference.

\subsection{Continual Domain Expansion (ConDo)}\label{sec:ConDo}
Due to the scene-dependent nature, the deployed APR model cannot generalize well to images that have highly different poses or scene conditions compared to the training data $\mathcal{S}^\Omega$. The key idea of Continual Domain Expansion (ConDo) is to continually update the APR model using the \emph{unlabeled} data seen naturally after model deployment, so that the model can generalize to more novel poses and scene conditions over time by simply running in the environment.

As shown in Fig.~\ref{Fig-pipeline}, ConDo starts from the model $\boldsymbol{\theta}$ trained on $(\mathcal{S}^\Omega, \mathcal{P}^\Omega)$. After model deployment, the (potentially multiple) clients, which use the newest model $\boldsymbol{\theta}$ to perform localization, upload (asyncronously) the observed images to the server. The server collects all newly received images at time step $k$. These images are added to the pool of unlabeled data $\Delta= \{\mathbf{I}_j^\Delta\}_{j=1}^M$ and the current model $\boldsymbol{\theta}_k$ is updated asynchronously from the clients using $\mathcal{S}^\Omega \bigcup \Delta$, with \emph{limited computation} that is much less than model re-training. $\boldsymbol{\theta}_k$ is re-deployed to the client after the update is finished. 

In the main experiment of Sec.~\ref{sec:exp}, we impose the constraint so that the compute for the pre-exectued model training plus all update rounds of ConDo is the same as training one APR model from scratch on $\mathcal{S}^\Omega \bigcup \Delta$. This ensures that each ConDo update round uses much less compute and time compared to model re-training, so that it can be applied to life-long scenarios. We also experiment with various fixed computation budgets to validate the effectiveness of ConDo in applications with different resource limits.

To expand the generalization domain of APR without forgetting, we \emph{uniformly} sample images from $\mathcal{S}^\Omega \bigcup \Delta$ to form a training batch during the model update. Though unsupervised domain adaptation (UDA) methods~\cite{chen2021representation, nejjar2023dare} have been proposed for standard image classification and regression, empirically (Sec.~\ref{sec:anal}) they are not effective for APR. To generate effective supervision on unlabeled data $\Delta$, we opt for a distillation-based approach. 
 Inspired by the fact that scene-independent methods~\citep{arandjelovic2016netvlad,sarlin2019coarse,von2022dm}, though slower and more memory consuming during inference, are much more robust than APR to novel poses and scene conditions, we distill the knowledge from these methods to APR using $\Delta$, so that the inference model can still maintain the memory and computation efficiency. Specifically, given a scene-independent method $f_\text{teacher}(\cdot)$, and a batch of data $\mathcal{B}^\Omega \bigcup \mathcal{B}^\Delta$ sampled from $\mathcal{S}^\Omega\bigcup\Delta$, the training objective is
 \begin{equation}\label{eq:obj}
 \small{
     \underset{\boldsymbol{\theta}}{\text{minimize}} \frac{\underset{\mathbf{I}^\Omega \in \mathcal{B}^\Omega}{\sum} \mathcal{L} (\mathbf{I}^\Omega, \mathbf{t}^*_{\mathbf{I}^\Omega}, \mathbf{r}^*_{\mathbf{I}^\Omega}) + \underset{\mathbf{I}^\Delta \in \mathcal{B}^\Delta}{\sum} \mathcal{L}_{\text{distill}} (\mathbf{I}^\Delta, f_{\text{teacher}})}{|\mathcal{B}^\Omega|+|\mathcal{B}^\Delta|},
     }
 \end{equation}
where $\mathbf{t}^*_{\mathbf{I}^\Omega}, \mathbf{r}^*_{\mathbf{I}^\Omega}$ are the ground truth pose of $\mathbf{I}^\Omega$. 
We choose HLoc~\citep{sarlin2019coarse} as the default $f_\text{teacher}$, with scene map built on $(\mathcal{S}^\Omega, \mathcal{P}^\Omega)$ (See Table.\ref{Tab-teacher-effects} for the robustness of ConDo with other teachers). We set $\mathcal{L}_\text{distill} = \mathcal{L}(\mathbf{I}^\Delta, f_{\text{teacher}}(\mathbf{I^\Delta}))$, i.e., substituting the output of $f_\text{teacher}$ into Eq.~\eqref{eq:loss}. As shown later in Sec.~\ref{sec:exp}, this simple yet effective loss is sufficient to approach the performance of training with ground truth poses on $\Delta$, and is robust to the choice of $f_\text{teacher}$. Meanwhile, distilling knowledge on data from new domains not only benefits the performance on the same domain, but can also improve the general robustness of APR, especially under scene condition changes.

By default, we assume the server has sufficient storage to maintain all historical data. For applications with limited server storage, we apply \emph{reservoir sampling}~\citep{rebuffi2017icarl} to update the replay buffer. Given a sequence of $K$ images and a storage $\mathcal{M}$ sufficient to maintain $N$ images, we push the first $N$ images to the storage as usual. For the $i$-th image where $i>N$, we generate a random integer $\alpha$ between $[1,i]$. If $\alpha \le N$, we replace the $\alpha$-th image stored in $\mathcal{M}$ with the $i$-th image in the sequence. Otherwise, we drop the $i$-th image. As shown later in Sec.~\ref{sec:anal}, ConDo with reservoir sampling is robust to the replay buffer size.

For architectures capable of handling multiple scenes~\citep{shavit2021learning,shavit2023coarse}, ConDo can also be applied when training data of new scenes and the inference data of old scenes arrive sequentially and interchangeably. When training data of a new scene is available during ConDo, we simply add an extra regression head to cope with new scene coordinates and perform normal ConDo training following Eq.~\eqref{eq:obj}. The only difference is that now the replay data of ConDo are sampled from all observed scenes. This strategy ensures that APR can handle both data from the same scene and the sequentially revealed multiple scenes, with a minor model parameter increase.

\section{Benchmark}\label{sec:benchmarks}
\begin{figure}
\centering
\includegraphics[width=\linewidth]{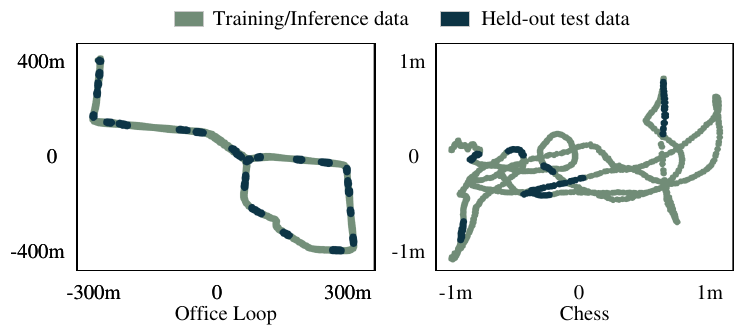}
\caption{Data split visualization. $\frac{1}{8}$ images in each scan (training and inference) are held out for evaluations. To create challenging evaluation data, We randomly hold several sets of images where each set is a continuous trajectory of the scan consisting of $16$ images. Left: Outdoor \emph{Office Loop} data. Right: Indoor \emph{Chess} scene in \emph{7Scenes}.\label{Fig-split-vis}
}
\end{figure}

To thoroughly evaluate ConDo in practical scenarios, we construct large-scale benchmarks covering both the change of \emph{scene conditions} (lighting, weather, season) and \emph{camera poses}. Specifically, we collect public datasets with multiple rounds of scans of the same scene. To simulate the practical scenario, we split multiple scans of the same scene into training and inference and reveal the inference scans sequentially, i.e., every round of ConDo model update starts when a new inference scan is revealed. We randomly hold out $\frac{1}{8}$ images in each scan (training and inference) and use them to evaluate the generalization of APR on the corresponding scan. To create challenging evaluation data, instead of holding out individual images uniformly distributed in each scan, we hold several sets of images where each set is a continuous trajectory of the scan consisting of $16$ images (see Fig.~\ref{Fig-split-vis}). The held-out evaluation data allow us to fully evaluate APR on images unseen both during normal training and ConDo.

To simulate \emph{novel poses and the sequentially revealed multiple scenes}, we adopt standard APR datasets, namely, \emph{7Scenes} and \emph{Cambridge}~\citep{glocker2013real,kendall2015posenet}. These two datasets represent the case of indoor and outdoor scenes respectively and different scans of the same scene contain distinct trajectories, which are suitable to evaluate the case of novel poses. We adopt the same training and inference split as in the baseline APR methods~\citep{kendall2015posenet,shavit2021learning}. Please refer to Appendix.~\ref{sec:benchmark-setting} for detailed information about the train/inference scan split, multi-scene revealing order, the used coordinate system, etc. 

\begin{figure}[!ht]
\centering
\includegraphics[width=\linewidth]{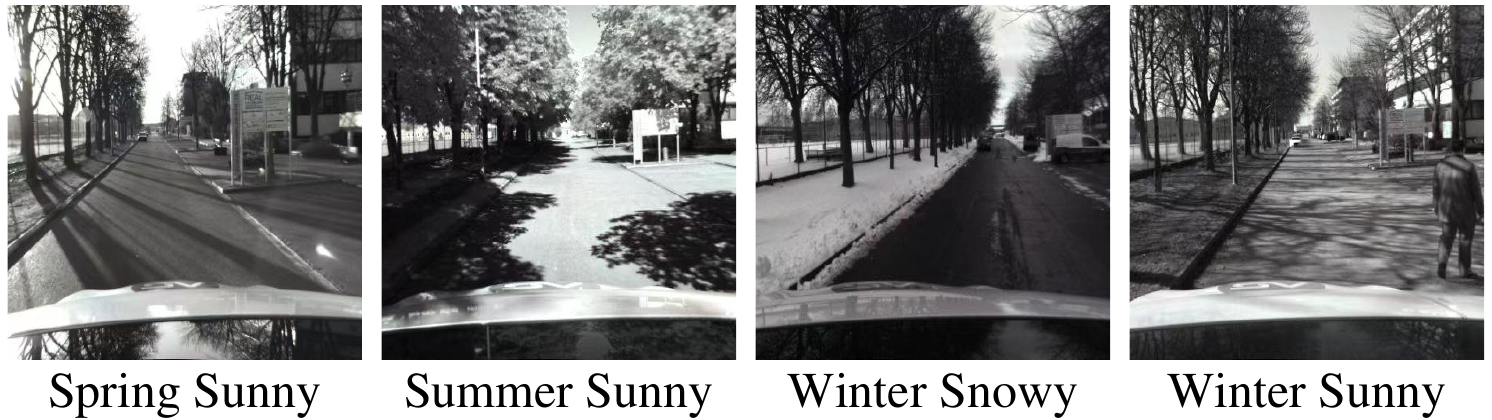}
\caption{\emph{Office Loop} images. Obvious differences exist between training (Spring Sunny) and inference scans, e.g., over-exposure (Summer Sunny), snow (Winter Snowy) and moving objects (Winter Sunny).}\label{Fig-samples}
\end{figure}

The drawbacks of \emph{7Scenes} and \emph{Cambridge} are the limited scene scale ($<140m \times 40m$) and scene condition change. The lighting and weather conditions of both datasets remain similar in different scans, and there are no obvious \emph{long-term scene changes (seasonal)} observed. To address these issues, we utilize large-scale driving datasets with both significant lighting changes (daytime to night time) and long-term scene changes (spring to winter). Specifically, we take the \emph{Office Loop} and \emph{Neighborhood}, which are two large-scale scenes in \emph{4Seasons}~\citep{wenzel2020fourseasons} with a sufficient amount of scans ($>6$) in the same scene. Each scan in these two scenes has a $>2km$ trajectory spanned at multiple city blocks, which is much larger than conventional APR datasets. See Fig.~\ref{Fig-samples} for sample images from different scene scans and Appendix.~\ref{sec:benchmark-setting} for the concrete train/inference scan split.


\section{Experiments}\label{sec:exp}
\noindent\textbf{Architectures.} We validate ConDo on two representative APR architectures, namely PoseNet (PN)~\citep{kendall2015posenet} and Pose-Transformer (PT)~\citep{shavit2021learning,shavit2023coarse}, which covers respectively the classic APR architectures for single and multiple scenes. 
\begin{table*}[!ht]
\centering
\resizebox{\linewidth}{!}{
\begin{tabular}{c|c|cccc|cccc}
\hline
\multirow{3}{*}{Model} & \multirow{3}{*}{Strategy} & \multicolumn{4}{c|}{\emph{Office Loop}}                                                            & \multicolumn{4}{c}{\emph{Neighborhood}}                                                           \\
                       &                           & \multicolumn{2}{c|}{Training scan held-out data} & \multicolumn{2}{c|}{Inference scan held-out data} & \multicolumn{2}{c|}{Training scan held-out data} & \multicolumn{2}{c}{Inference scan held-out data} \\
                       &                           & \multicolumn{1}{c}{Median}               & \multicolumn{1}{c|}{Mean}                 & Median               & \multicolumn{1}{c|}{Mean}                  & Median               & \multicolumn{1}{c|}{Mean}                 & Median               & \multicolumn{1}{c}{Mean}                  \\
\hline
\multirow{4}{*}{PN}    & 1.Train-only                & 2.03/0.60            & \multicolumn{1}{c|}{2.54/1.08}            & 18.10/2.68           & 100.42/16.42          & 1.19/0.79            & \multicolumn{1}{c|}{1.57/1.33}            & 10.99/3.74           & 27.67/13.88           \\
                       & 2.ConDo                     & 1.66/0.25            & \multicolumn{1}{c|}{2.03/0.37}            & 2.16/0.52            & 2.61/0.88             & 1.12/0.33            & \multicolumn{1}{c|}{1.36/0.68}            & 1.14/0.45            & 1.39/0.59             \\
                       & 3.Re-train with GT          & 1.64/0.20            & \multicolumn{1}{c|}{2.05/0.30}            & 1.80/0.19            & 2.42/0.51             & 0.99/0.26            & \multicolumn{1}{c|}{1.23/0.37}            & 1.00/0.23            & 1.29/0.31             \\
                       & \cellcolor{gray!40}Improvement (1$-$2)            & \cellcolor{gray!40}\textbf{0.37/0.35}            & \multicolumn{1}{c|}{\cellcolor{gray!40}\textbf{0.51/0.71}}           & \cellcolor{gray!40}\textbf{15.94/2.16}           & \cellcolor{gray!40}\textbf{97.81/15.54}            & \cellcolor{gray!40}\textbf{0.07/0.46}            & \multicolumn{1}{c|}{\cellcolor{gray!40}\textbf{0.21/0.65}}           & \cellcolor{gray!40}\textbf{9.85/3.29}           & \cellcolor{gray!40}\textbf{26.28/13.29} \\
\hline
\multirow{4}{*}{PT}    & 1.Train-only                & 1.70/0.29            & \multicolumn{1}{c|}{1.82/0.84}            & 6.12/16.14           & \multicolumn{1}{c|}{42.15/43.24}           & 1.22/0.35            & \multicolumn{1}{c|}{1.33/0.75}            & 2.99/1.33            & 17.69/23.64        \\
                       & 2.ConDo                     & 1.34/0.21            & \multicolumn{1}{c|}{1.45/0.49}            & 1.50/0.49            & 1.86/1.31             & 0.87/0.24            & \multicolumn{1}{c|}{0.94/0.40}            & 0.89/0.38            & 1.04/0.50             \\
                       & 3.Re-train with GT          & 1.41/0.19            & \multicolumn{1}{c|}{1.39/0.64}            & 1.46/0.18            & 1.59/0.63             & 0.73/0.22            & \multicolumn{1}{c|}{0.77/0.36}            & 0.76/0.19            & 0.84/0.33             \\ 
                       & \cellcolor{gray!40}Improvement (1$-$2)& \cellcolor{gray!40}\textbf{0.36/0.08}            & \multicolumn{1}{c|}{\cellcolor{gray!40}\textbf{0.37/0.35}}           & \cellcolor{gray!40}\textbf{4.62/15.65}           & \cellcolor{gray!40}\textbf{40.29/41.93}            & \cellcolor{gray!40}\textbf{0.35/0.11}            & \multicolumn{1}{c|}{\cellcolor{gray!40}\textbf{0.39/0.35}}           & \cellcolor{gray!40}\textbf{2.10/0.95}           & \cellcolor{gray!40}\textbf{17.65/23.14} \\
\hline
\end{tabular}
}
\caption{Results on scene condition changes. Position ($m$) / orientation ($^\circ$) errors are reported for various strategies and architectures. Results on training and inference scans are reported separately for better analysis. ConDo improved the deployed APR (\emph{Train-only}) by a large margin across architectures and scenes, reaching near upper bound \emph{Re-train with GT} performance with limited computation budgets for model updates.}\label{Tab-condition}
\end{table*}

\begin{table*}[!ht]
\centering
\resizebox{\linewidth}{!}{
\begin{tabular}{c|c|cccc|cccc}
\hline
\multirow{3}{*}{Model} & \multirow{3}{*}{Strategy} & \multicolumn{4}{c|}{\emph{7Scenes}}                                                            & \multicolumn{4}{c}{\emph{Cambridge}}                                                           \\
                       &                           & \multicolumn{2}{c|}{Training scan held-out data} & \multicolumn{2}{c|}{Inference scan held-out data} & \multicolumn{2}{c|}{Training scan held-out data} & \multicolumn{2}{c}{Inference scan held-out data} \\
                       &                           & \multicolumn{1}{c}{Median}               & \multicolumn{1}{c|}{Mean}                 & Median               & \multicolumn{1}{c|}{Mean}                  & Median               & \multicolumn{1}{c|}{Mean}                 & Median               & \multicolumn{1}{c}{Mean}                  \\
\hline
\multirow{4}{*}{PN}    & 1.Train-only                & 0.023/0.925          & \multicolumn{1}{c|}{0.027/1.236}          & 0.303/9.536          & 0.374/15.228          & 0.933/2.722          & \multicolumn{1}{c|}{1.284/4.889}          & 1.405/3.292           & 2.280/4.778          \\
                       & 2.ConDo                     & 0.069/2.198          &  \multicolumn{1}{c|}{0.078/2.460}        &   0.080/2.641                  & 0.097/3.042          &  1.259/3.131       &    \multicolumn{1}{c|}{1.711/5.573}    &   1.052/2.612        &  1.518/3.560         \\
                       & 3.Re-train with GT          & 0.024/0.995          & \multicolumn{1}{c|}{0.029/1.265}          & 0.024/1.016          & 0.028/1.191           & 0.786/2.446          & \multicolumn{1}{c|}{1.228/4.896}          & 0.864/2.155           & 1.288/2.797          \\
                       & \cellcolor{gray!40}Improvement (1$-$2)            & \cellcolor{gray!40}-0.046/-1.273            & \multicolumn{1}{c|}{\cellcolor{gray!40}-0.051/-1.224}           & \cellcolor{gray!40}\textbf{0.223/6.895}           & \cellcolor{gray!40}\textbf{0.277/12.186}            & \cellcolor{gray!40}-0.326/-0.409            & \multicolumn{1}{c|}{\cellcolor{gray!40}-0.427/-0.684}           & \cellcolor{gray!40}\textbf{0.353/0.680}           & \cellcolor{gray!40}\textbf{0.762/1.218}\\
\hline
\multirow{4}{*}{PT}    & 1.Train-only                & 0.021/1.158          & \multicolumn{1}{c|}{0.025/1.547}          & 0.198/8.373          & 0.284/12.162          & 0.806/2.502          & \multicolumn{1}{c|}{1.084/5.756}          & 1.101/2.682           & 1.877/3.536        \\
                       & 2.ConDo                     & 0.049/1.581          & \multicolumn{1}{c|}{0.054/1.885}          & 0.065/2.113          & 0.077/2.381           & 0.816/2.555          & \multicolumn{1}{c|}{1.170/6.608}          & 0.696/2.106           & 1.134/3.470          \\
                       & 3.Re-train with GT          &0.026/1.267          & \multicolumn{1}{c|}{0.030/1.535}        &  0.026/1.261         &  0.030/1.507   & 0.709/2.228          & \multicolumn{1}{c|}{1.041/4.738}          & 0.690/2.012           & 1.030/2.550          \\
                       & \cellcolor{gray!40}Improvement (1$-$2)            & \cellcolor{gray!40}-0.028/-0.423            & \multicolumn{1}{c|}{\cellcolor{gray!40}-0.029/-0.338}           & \cellcolor{gray!40}\textbf{0.133/6.260}           & \cellcolor{gray!40}\textbf{0.207/9.781}            & \cellcolor{gray!40}-0.010/-0.053            & \multicolumn{1}{c|}{\cellcolor{gray!40}-0.086/-0.852}           & \cellcolor{gray!40}\textbf{0.405/0.576}           & \cellcolor{gray!40}\textbf{0.743/0.066}\\
\hline
\end{tabular}
}
\caption{Results on pose changes. Position ($m$) / orientation ($^\circ$) errors are reported for various strategies and architectures. Though seeing data with novel poses and from other scenes might not always benefit the performance on historical data, ConDo still consistently and significantly improved the generalization on inference scans.}\label{Tab-pose}
\end{table*}
\noindent\textbf{Implementation.} Unless otherwise stated, the code and hyper-parameter settings of the baselines strictly follow the official code release. The original Pose-Transformer can use multiple regression heads and scene-dependent latent embeddings to handle multiple scenes. We only apply multiple regression heads since it is sufficient to achieve similar performance (Appendix~\ref{sec:ms-design}). APRs are first learned on training data until converging in the initial training. In the main experiment, we follow the setup of large scale continual learning~\cite{cai2021online} and limit the computation budget of ConDo by first identifying the budget $b = \text{epoch} * \text{iteration\_per\_epoch} * \text{batch\_size} / |\mathcal{S}^\Omega|$ for the baseline APR model to converge on the initial training data $\mathcal{S}^\Omega$. $b$ represents the average number of iterations required per image. Then for every round of ConDo update with $N$ images newly revealed, we assign $N*b/\text{batch\_size}$ training iterations (see Appendix~\ref{sec:benchmark-setting} for actual numbers of $b$) with the same batch size as the initial training, so that the whole ConDo procedure including initial training and all ConDo updates, consumes roughly only the budget to train one APR model from scratch on all revealed data. This ensures that we use \emph{much less} computation than model re-training in every round of ConDo update. See Sec.~\ref{sec:anal} for the comparison of ConDo and model re-training with varied computation budgets. All models are trained using one RTX-4090 GPU. 

\noindent\textbf{Evaluation protocol.} Following the standard~\citep{kendall2015posenet}, we compute the median/mean camera position ($m$) and orientation (${^\circ}$) error for different methods. For baselines, we train the model on the initial training data and evaluate the performance on all held-out test data (from both the training and inference scans). For ConDo, we first train the model over the initial training data, perform ConDo updates sequentially on all inference scans, and then evaluate the final model on the held-out data.

\subsection{Main Results}\label{sec:main}
Table~\ref{Tab-condition} and~\ref{Tab-pose} show the main results on benchmarks constructed in Sec.~\ref{sec:benchmarks} with respectively the scene condition (\emph{Office Loop} and \emph{Neighbourhood}) and pose (\emph{7Scenes} and \emph{Cambridge}) changes. For each architecture (PN and PT), we show the results of 3 training frameworks: 
\setlist[enumerate]{label=\textbf{\arabic*.},itemsep=0pt,leftmargin=*,partopsep=0pt}
\begin{enumerate}
    \item \emph{Train Only}: Normal APR training on the initial data $\mathcal{S}^\Omega$. Representing the practical base APR performance.
    \item \emph{ConDo}: The proposed ConDo strategy.
    \item \emph{Re-train with GT}: Train an APR model from scratch until convergence (\emph{infinite computation budget} \emph{at any time}) on both the training and inference data ($\mathcal{S}^\Omega \bigcup \Delta$), \emph{with the GT} label on $\Delta$ provided. This setup estimates the best performance that ConDo can achieve.
\end{enumerate}
See Sec.~\ref{sec:anal} for further comparisons between ConDo and standard UDA methods.

\begin{figure*}[!ht]
    \centering
    \includegraphics[width=0.85\linewidth]{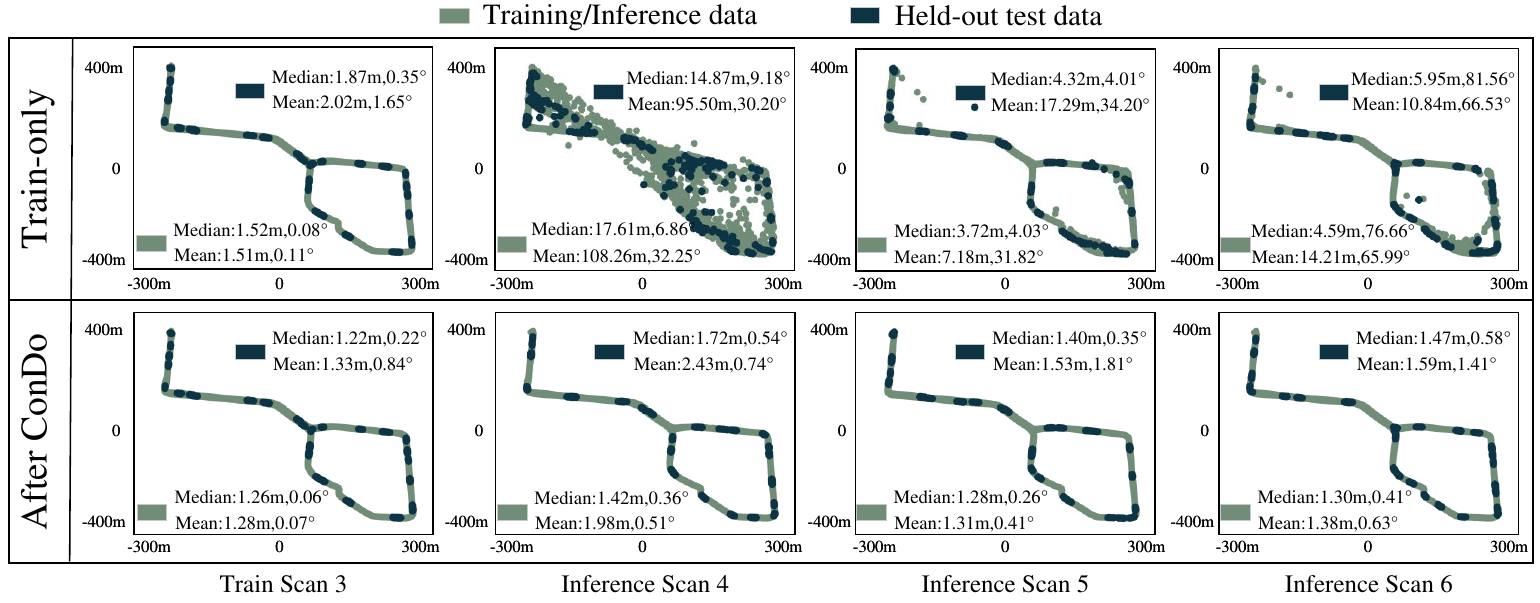}
    \caption{Result visualization on \emph{Office Loop}. We visualize results on training and inference scans, where dark blue points indicate held-out test data and grey-green indicates training/inference data. Due to the space limit, we only visualize one training scan (\emph{Train Scan 3}), see Appendix~\ref{sec:more-traj} for other training scans. \emph{Train-only} performed well on \emph{Train Scan 3}, but cannot handle unseen scene condition changes (top row). By updating with unlabelled inference data, ConDo not only adapted to inference scans, but also generalized to the training ones (1.87m to 1.22m on the held-out data of \emph{Train Scan 3}).}
    \label{Fig-scan-results}
\end{figure*}

\emph{ConDo} significantly improved the performance across baseline architectures and datasets. This shows the capability of ConDo to adapt to both scene condition change (\emph{Office Loop} and \emph{Neighbourhood}) and data from novel poses (\emph{7Scenes} and \emph{Cambridge}), and sequentially learn to localize in multiple scenes (\emph{7Scenes} and \emph{Cambridge}). Before ConDo, the mean error of the baselines was much larger than the median error on the inference scans of large-scale datasets. E.g.,  PT had $42.15m$ mean error vs $6.12m$ median error on \emph{Office Loop}. This indicates the existence of catastrophically failing predictions (see Fig.~\ref{Fig-scan-results} for visualizations). After ConDo, not only mean and median errors were reduced significantly (by $23$x and $4$x respectively), but also the difference between them became small. This change shows the significantly improved generalization of ConDo. 

The performance difference between \emph{ConDo} and \emph{Re-train with GT} was small, even though 1) ConDo only used unlabeled inference data, and 2) used limited compute for model updates on sequentially revealed data. For example, \emph{Re-train with GT} on Office Loop with PT used $\sim120h$ to reach the reported performance and performed much worse with less compute (see Sec.~\ref{sec:anal}), while each ConDo update round only took $\sim20h$, i.e., achieving similar accuracy with only $\frac{1}{6}$ of the compute. Note that this difference will further increase over time with more data collected. Interestingly, \emph{ConDo} performed marginally better \emph{Re-train with GT} in \emph{Office Loop}, it was because HLoc we used is very accurate in this dataset, especially in terms of translation (see Table.~\ref{Tab-teacher-effects} for details).

Another interesting observation is that whether new data helps the general robustness of an APR model depends on the type of change in the data. For data with scene condition changes (Table~\ref{Tab-condition}), learning from more data significantly improved the accuracy not only on the new scans, but also on previously seen scans. This result shows that training on more diverse data with different scene conditions helps to improve the general robustness of APR. On data with pose changes (Table~\ref{Tab-pose}), seeing new data may not always help the performance of old ones, even with ground truth labels (\emph{Re-train with GT}). Appendix.~\ref{sec:multi_scene_performance_drop} further analyzes the detailed reason for this phenomenon.

\begin{figure}[!ht]
    \centering
    \includegraphics[width=0.95\linewidth]{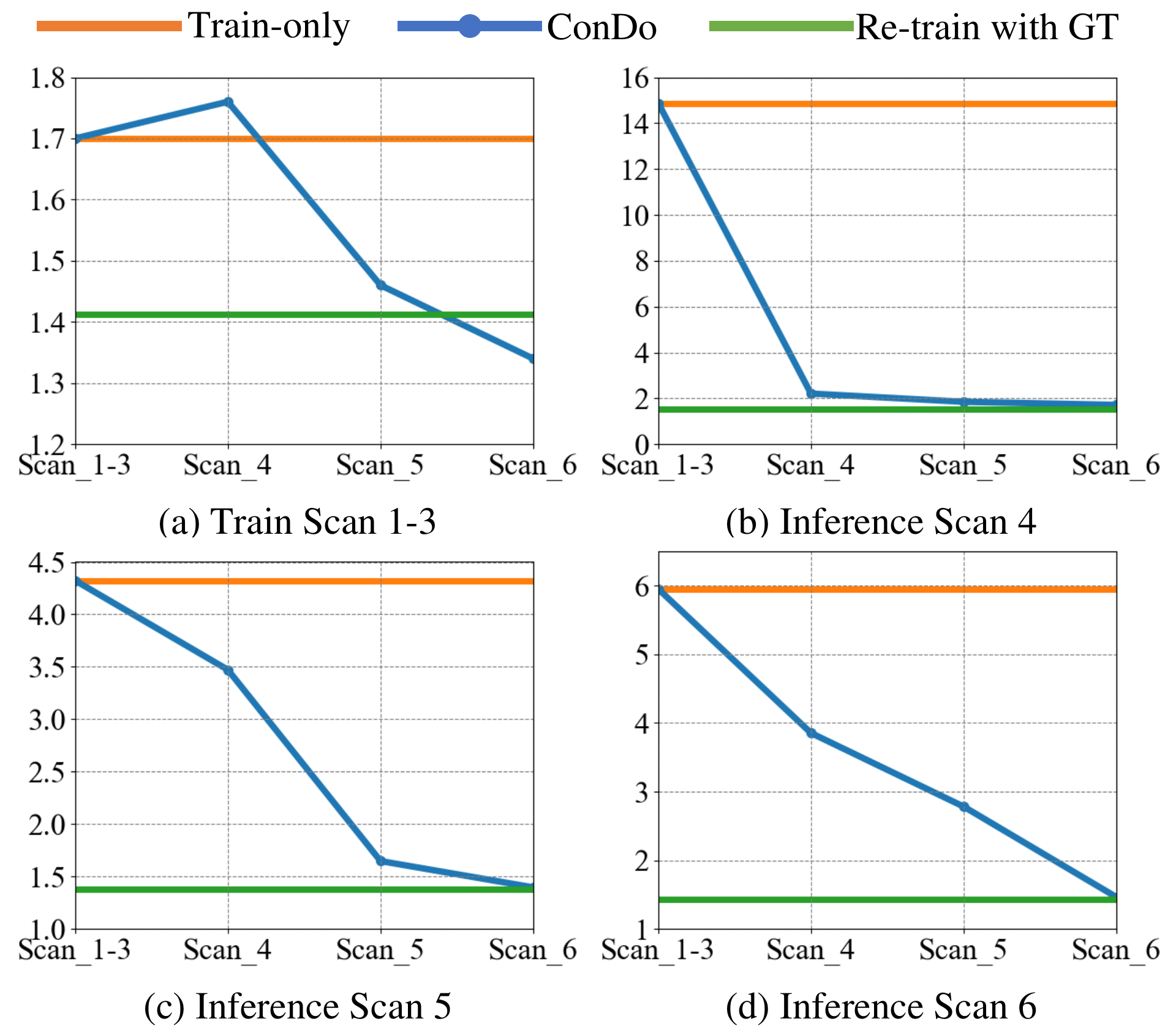}
    \caption{Intermediate ConDo performance. The median position error ($m$) of PT on \emph{Office Loop} is reported. The x axis indicates the scans seen in each round of ConDo update. ConDo updates improved the accuracy not only on the current scan, but also on other scans. See Appendix~\ref{sec:rounds} for the version with \emph{Train Scan 1-3} in separate figures.}
    \label{Fig-line-chart}
\end{figure}

Fig.~\ref{Fig-scan-results} visualizes the result of PT on \emph{Office Loop}. Consistent with the conclusion from quantitative results, due to the weather/lighting condition change, the performance of APR dropped significantly on the inference scans, with many severe localization errors, especially in \emph{Inference Scan 4}. After ConDo, these severe errors completely disappeared, and the localization accuracy not only improved on the inference scans, but also on the training ones (1.87m to 1.22m on the held-out data \emph{Train Scan 3}). Fig.~\ref{Fig-line-chart} shows the model accuracy on the held-out data of different scans after each round of ConDo update, which shows that seeing more data during ConDo can improve the \emph{general robustness} of APR, resulting in a steady accuracy improvement. Note that the accuracy improvement on the training data was not caused by more training iterations in ConDo, since we have ensured that the model has converged during the \emph{Train Only} phase, i.e., more training iterations without additional data would not help. 

\subsection{Analysis} \label{sec:anal}
This section analyzes the effectiveness of individual ConDo components, and we report results on \emph{Office Loop} in the format of position ($m$)/orientation ($^\circ$) error. 
\begin{table}[!ht]
\centering
\resizebox{\linewidth}{!}{
\begin{tabular}{c|cc|cc}
\hline
\multirow{2}{*}{Strategy}  & \multicolumn{2}{c|}{Training scan held-out data} & \multicolumn{2}{c}{Inference scan held-out data}  \\
        		     & Median               & Mean                 & Median               & Mean    \\
\hline
Train-only$_{512}$           & 2.00/0.29   & 2.20/0.74   & 6.79/12.77  & 55.28/44.41  \\
RSD        & 1.99/0.30   & 2.09/0.80   & 6.65/12.52  & 53.55/43.30 \\
DARE             & 1.86/0.29   & 1.98/0.73   & 6.72/13.08  & 54.19/44.13  \\
\hline
Train-only         & 1.70/0.29   & 1.82/0.84  & 6.12/16.14 & 42.15/43.24  \\
MIC$_{mae}$                       & 2.59/0.73  & 3.03/2.19   & 6.29/39.87  & 20.03/50.99  \\
MIC$_{moco}$                      & 2.30/0.54   & 4.51/3.86   & 5.74/77.61  & 15.37/63.77  \\
MIC$_{mae}$+ConDo             & 2.42/0.69   & 2.67/1.81   & 4.39/4.30   & 8.42/18.85   \\
MIC$_{moco}$+ConDo             & 2.19/0.45   & 4.34/1.36   & 4.15/2.62   & 10.19/19.78  \\
ConDo                      & \textbf{1.34/0.21}   & \textbf{1.45/0.49}   & \textbf{1.50/0.49}   & \textbf{1.86/1.31}    \\
\hline
\end{tabular}
}
\caption{Comparison to UDA methods. We compare ConDo with RSD and DARE which are the latest UDA regression strategies, as well as the most effective and applicable strategy in UDA classification, MIC, and we modified its output from class distribution to regression space for the localization task.}\label{Tab-UDA-comparison}
\end{table}
\begin{table}[!th]
\resizebox{\linewidth}{!}{
\centering
\begin{tabular}{c|c|c|cc|cc}
\hline
\multirow{2}{*}{Rate} & \multirow{2}{*}{Time} & \multirow{2}{*}{Strategy} & \multicolumn{2}{c|}{Train scan held-out data} & \multicolumn{2}{c}{Infer scan held-out data} \\
                              &                             &                           & Median                  & Mean                    & Median                  & Mean                     \\ \hline
unlimited           & 120h        & Re-train         &  1.41/0.19 &  1.39/0.64 &  1.46/0.18  &  1.59/0.63   \\ 
\hline
\multirow{2}{*}{1}           & \multirow{2}{*}{20h}        & Re-train                  & 1.86/0.24                 & 1.90/0.65                  &  1.79/0.44                 &  2.01/1.19                  \\
                              &                             & \cellcolor{gray!40}ConDo                     & \cellcolor{gray!40}1.34/0.21               & \cellcolor{gray!40}1.45/0.49               & \cellcolor{gray!40}1.50/0.49               & \cellcolor{gray!40}1.86/1.31                \\
\hline
\multirow{2}{*}{1/2}         & \multirow{2}{*}{10h}        & Re-train                  & 2.03/0.36               & 2.07/0.84               & 2.12/0.57               & 2.30/1.47                \\
                              &                             & \cellcolor{gray!40}ConDo                     & \cellcolor{gray!40}1.56/0.24                 & \cellcolor{gray!40}1.72/0.61                   & \cellcolor{gray!40}1.64/0.50                 &   \cellcolor{gray!40}1.92/1.39                 \\
\hline
\multirow{2}{*}{1/4}         & \multirow{2}{*}{5h}         & Re-train                  & 2.75/0.61               & 2.97/1.20               & 2.70/0.65               & 3.11/1.58                \\
                             &                             & \cellcolor{gray!40}ConDo                     & \cellcolor{gray!40}1.81/0.27               & \cellcolor{gray!40}1.88/0.64               & \cellcolor{gray!40}1.91/0.52               & \cellcolor{gray!40}2.34/1.49                \\
\hline
\multirow{2}{*}{1/8}         & \multirow{2}{*}{2.5h}       & Re-train                  & 3.46/0.78               & 3.64/1.56               & 3.42/0.87               & 3.68/1.98                \\
                              &                             & \cellcolor{gray!40}ConDo                     & \cellcolor{gray!40}1.95/0.35               & \cellcolor{gray!40}2.03/0.69               & \cellcolor{gray!40}2.23/0.62               & \cellcolor{gray!40}2.51/1.64                \\
\hline
\multirow{2}{*}{1/100}       & \multirow{2}{*}{12min}      & Re-train                  & 8.13/2.28               & 9.38/7.36               & 9.12/2.62               & 11.49/8.03               \\
                              &                             & \cellcolor{gray!40}ConDo                     & \cellcolor{gray!40}2.41/0.47               & \cellcolor{gray!40}2.42/2.53               & \cellcolor{gray!40}2.89/0.91               & \cellcolor{gray!40}3.31/2.80                \\ \hline
\end{tabular}
}
\caption{ConDo vs Re-train with varied training budgets. ConDo reached a similar accuracy up to 25x faster than Re-train.}\label{Tab-budget}
\end{table}

 \noindent\textbf{ConDo vs UDA.} As mentioned in Sec.~\ref{sec:ConDo}, unsupervised domain adaptation (UDA) is widely used to adapt models to novel data. In Table.~\ref{Tab-UDA-comparison}, we compare ConDo with 3 most applicable UDA baselines, namely, \emph{RSD}~\citep{chen2021representation}, \emph{DARE}~\citep{nejjar2023dare} and \emph{MIC$_{mae}$}~\citep{hoyer2023mic}. Following the original papers, we reduce the feature dimension of RSD and DARE (from $1024$) to $512$ to avoid divergence. Empirically, RSD still diverges with this strategy, hence we report its result before divergence. We also run \emph{Train-only$_{512}$} with $512$ dimension features to show the improvements of RSD and DARE. We also try \emph{MIC$_{moco}$} which replaces masked inputs of MIC$_{mae}$ with augmented ones. Since MIC is compatible with ConDo supervision, we also combine it with ConDo (MIC$_{mae/moco}$+ConDo). As shown in Table.~\ref{Tab-UDA-comparison}, RSD and DARE have minor improvements over Train-only$_{512}$ and are far behind ConDo. MIC$_{mae}$ and MIC$_{moco}$ hurt both Train-only and ConDo.

 \noindent\textbf{Computation budget.} Practical applications have varied computation budgets. To demonstrate the effectiveness of ConDo, we compare it with \emph{Re-train with GT} (Re-train in short) under the same budget limits. Note that the \emph{Re-train with GT} in the main results used unlimited compute, and took $120h$ for the last round of update, 6x higher than the ConDo update. As shown in Table.\ref{Tab-budget}, we gradually reduce the budget limit from $20h$ (the same as ConDo in the main results) to $12min$ ($\frac{1}{100}$ of the original budget). ConDo reached a similar accuracy much faster than \emph{Re-train with GT}, even without using GT. E.g., the performance of ConDo with just $12min$ of model updates was on-par with \emph{Re-train with GT} for $5h$ --- a 25x compute/time reduction. Note that with only $20h$, \emph{Re-train} performed much worse than ConDo, even with GT. Appendix \ref{sec:total-time} further shows the accumulated time of ConDo and \emph{Re-train with GT} after all update rounds.
\begin{figure}[!th]
    \centering
    \includegraphics[width=0.85\linewidth]{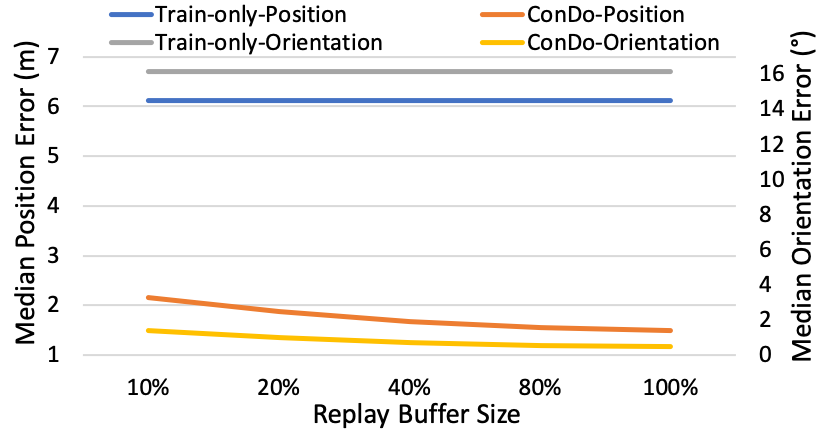}
    \caption{Effect of replay buffer sizes. The horizontal axis is the ratio between the replay buffer size and the whole dataset size. The vertical axis reports median errors in \emph{Office Loop}.}
    \label{Fig-replay-buffer}
\end{figure}
\begin{table}[!ht]
\centering
 \resizebox{\linewidth}{!}{
\begin{tabular}{c|cc|cc|cc}
\hline
\multirow{2}{*}{$f_{teacher}(\cdot)$}  & \multicolumn{2}{c|}{Train held-out data} & \multicolumn{2}{c|}{Infer held-out data} & \multicolumn{2}{c}{Teacher err in infer scan}  \\
                 & Median               & Mean                 & Median               & Mean    & Median & Mean \\
\hline
DM-VIO                 & 1.50/0.29           & 1.67/0.55           & 5.21/0.81         & 5.56/1.17    & 5.08/0.82 & 5.74/0.84         \\
ORB-SLAM                    & 1.40/0.20            & 1.43/0.42           & 3.39/0.94            & 3.63/1.24    & 3.15/0.93  & 3.14/0.97       \\
NetVLAD                & 1.88/0.33            & 2.00/0.83           & 3.00/1.02          & 49.39/10.21       & 0.97/1.13 & 44.50/10.61       \\
HLoc                     & 1.34/0.21            & 1.45/0.49            & 1.50/0.49            & 1.86/1.31   & 0.05/0.31 & 0.41/0.51          \\
GT                       & 1.46/0.20            & 1.47/0.43            & 1.58/0.19            & 1.66/0.69   & - & -           \\      
\hline
\end{tabular}
}
\caption{Effect of teacher models.  Results are evaluated on Office Loop and reported in the form of position ($m$) / orientation ($^\circ$) errors. Left and middle column: results on held-out evaluation data of ConDos supervised by different teachers. Right column: teacher performance on the \emph{whole set of} inference scans. Though slightly worse than HLoc, all other teachers can be effectively applied to ConDo and provide a reasonable performance improvement.}\label{Tab-teacher-effects}
\end{table}

\noindent\textbf{Replay buffer size.} For applications with limited server storage, one can apply the reservoir buffer during ConDo. Following the standard approach~\citep{cai2023clnerf}, we apply \emph{reservoir sampling} to update ConDo and analyze its performance under different replay buffer sizes. As shown in Fig.~\ref{Fig-replay-buffer}, the performance ConDo only dropped slightly even under extreme replay buffer size limitation ($10\%$ of the overall dataset size), which still significantly improved over the \emph{Train-only} baseline. This shows the effectiveness of ConDo in practical applications with strong storage limits.

\noindent\textbf{Teacher.} Table.\ref{Tab-teacher-effects} shows the performance of ConDo with different teachers $f_\text{teacher}(\cdot)$ (see Appendix~\ref{sec:teacher-performance} for implementation details). The right-most column also reports the prediction error of each teacher model. The accuracy of ConDo and the teacher is positively correlated. On the other hand, despite the higher prediction noise, ConDo with other teachers all provided reasonable improvements to the base APR model.

\noindent\textbf{Pre-training.} Stronger pre-trained backbones often improve model generalization~\citep{keetha2023anyloc,kappeler2023few}. Appendix.\ref{sec:dino} demonstrates that replacing the original APR backbones with stronger ones~\citep{oquab2023dinov2} cannot replace the functionality of ConDo.

\section{Conclusion}\label{sec:conclusion}
We have identified the problem of APR in generalizing to novel data during inference. We have proposed Continual Domain Expansion (ConDo) to address this problem. By distilling knowledge from scene-independent localization methods, ConDo allows APR to improve steadily and continually while running in deployed environments with unlabeled data. We have constructed large-scale benchmarks covering 1) indoor and outdoor scenes, and 2)  the change of both environment conditions and novel poses. Experiments have verified the effectiveness and robustness of ConDo under varied teacher models, model architectures, scene types, compute budgets and replay buffer sizes.

\bibliography{aaai25}

\clearpage
\appendix

\section{Appendix}
\subsection{Inference Memory and Time Cost}\label{sec:memory-time}
\begin{table}[!h]
\centering
\resizebox{.9\linewidth}{!}{
\begin{tabular}{cccc}
\toprule
Methods & Type & Time  & Memory (10 scenes)  \\
\midrule
ActiveSearch     & Optimization  & 375ms & TB            \\
NetVlad & Retrieval & 13ms  & GB            \\
HLoc   & Retrieval+Optimization & 73ms  & GB           \\
DSAC$^*$   & Coordinate Regression & 30ms  & GB          \\
PoseNet & APR  & 8ms & MB        \\
PoseTransformer & APR & 12ms  & MB     \\
\bottomrule
\end{tabular}
}
\caption{Inference memory and time costs of visual localization methods.}\label{Tab-time-memory-cost}
\end{table}
Though less robust than scene-independent methods, APR is much more memory and time efficient. Following ~\citep{shavit2023coarse}, Table.~\ref{Tab-time-memory-cost} below shows the runtime and memory cost of representative visual localization methods. APR is the only one that achieves $\sim10ms$ inference time and MB-level memory. Also, APR does not store reference images, which has fewer privacy constraints.

\subsection{Benchmark Settings}\label{sec:benchmark-setting}
\begin{table}[!htp]
\centering
\resizebox{\linewidth}{!}{
\begin{tabular}{cccc}
\toprule
Dataset                                                                                       & Split                                          & Scan  & Tags                    \\ 
\midrule
\multicolumn{1}{c|}{\multirow{6}{*}{\vspace{-11pt}\begin{tabular}[c]{@{}c@{}}\emph{Office}\\ \emph{Loop}\end{tabular}}}   & \multirow{3}{*}{\vspace{-2pt}Train}                     & 1 & spring,sunny,afternoon  \\
\multicolumn{1}{l|}{}                                                                         &                                                & 2 & spring, sunny, afternoon  \\
\multicolumn{1}{l|}{}                                                                         &                                                & 3 & spring, sunny, morning    \\ \cmidrule(lr){2-4} 
\multicolumn{1}{l|}{}                                                                         & \multicolumn{1}{l}{\multirow{3}{*}{\vspace{-2pt}Inference}} & 4 & summer, sunny, morning    \\ 
\multicolumn{1}{l|}{}                                                                         & \multicolumn{1}{l}{}                           & 5 & winter, snowy, afternoon  \\
\multicolumn{1}{l|}{}                                                                         & \multicolumn{1}{l}{}                           & 6 & winter, sunny, afternoon  \\ 
\midrule
\multicolumn{1}{c|}{\multirow{7}{*}{\vspace{-11pt}\begin{tabular}[c]{@{}c@{}}\emph{Neighbor}\\ \emph{hood}\end{tabular}}} & \multirow{3}{*}{\vspace{-2pt}Train}                     & 1 & spring, cloudy, afternoon \\
\multicolumn{1}{c|}{}                                                                         &                                                & 2 & fall, cloudy, afternoon   \\
\multicolumn{1}{c|}{}                                                                         &                                                & 3 & fall, rainy, afternoon    \\ \cmidrule(lr){2-4}
\multicolumn{1}{c|}{}                                                                         & \multirow{4}{*}{\vspace{-3pt} Inference}                     & 4 & winter, cloudy, morning   \\
\multicolumn{1}{c|}{}                                                                         &                                                & 5 & winter, sunny, afternoon  \\
\multicolumn{1}{c|}{}                                                                         &                                                & 6 & spring, cloudy, evening   \\
\multicolumn{1}{c|}{}                                                                         &                                                & 7 & spring, cloudy, evening  \\
\bottomrule
\end{tabular}
}
\caption{Training and inference scan splits in \emph{Office Loop} and \emph{Neighbourhood}. Tags show short-term and long-term scene changes (lightning, weather, season).}\label{Tab-condition-split}
\end{table}
\begin{table}[!h]
\centering
\resizebox{\linewidth}{!}{
\begin{tabular}{cccc}
\toprule
Dataset                                                         & Scene                          & Train Scans                                                                                 & Inference Scans                     \\ 
\midrule
\multicolumn{1}{c|}{\multirow{7}{*}{\emph{7Scenes}}}    & Chess         & 01,02,04,06          &03,05       \\
\multicolumn{1}{c|}{}                 & Fire          & 01,02                        & 03,04      \\
\multicolumn{1}{c|}{}                & Heads         & 02                                                & 01          \\
\multicolumn{1}{c|}{}               & Office        & 01,03,04,05,08,10             & 02,06,07,09 \\
\multicolumn{1}{c|}{}              & Pumpkin       & 02,03,06,08         & 01,07       \\ \cmidrule(lr){2-4} 
\multicolumn{1}{c|}{}              & Redkitchen           & 01,02,05,07,08,11,13        & 03,04,06,12,14               \\
\multicolumn{1}{c|}{}             & Stairs        & 02,03,05,06                      & 01,04       \\
\midrule
\multicolumn{1}{c|}{\multirow{4}{*}{\emph{Cambridge}}}  & KingsCollege  & 01,04,05,06,08                       & 02,03,07    \\
\multicolumn{1}{c|}{}                  & OldHospital   & 01,02,03,05,06,07,09        & 04,08       \\ \cmidrule(lr){2-4} 
\multicolumn{1}{c|}{}                      & ShopFacade   & 02                & 01,03       \\
\multicolumn{1}{c|}{}                 & StMarysChurch & 01,02,04,06,07,08,09,10,12,14  & 03,05,13    \\
\bottomrule
\end{tabular}
}
\caption{Multi-scene splits}\label{Tab-scene-split}
\end{table}
This section shows more details of benchmark settings in Sec.~\ref{sec:benchmarks}.
Table.~\ref{Tab-condition-split} shows scan splits of \emph{Office Loop} and \emph{Neighbourhood} for training/inference, and the inference scans are reveal sequentially, i.e., every round of ConDo model update starts when a new inference scan is revealed.
 Table.~\ref{Tab-scene-split} shows the train-inference scan splits of \emph{7Scenes} and \emph{Cambridge}. We use default training and testing trajectories in the \emph{7Scenes}~\citep{glocker2013real} and \emph{Cambridge}~\citep{kendall2015posenet} as our training and inference scans. For the multi-scene revealing order, APRs are trained on training scans of 4 (\emph{7Scenes}) or 2 (\emph{Cambridge}) scenes in initial training and updated with inference scans of the same scenes, then expanded sequentially to two other scenes following the row order of Table.~\ref{Tab-scene-split} and inter-changeably with their training scans and inference scans. We set $b=4200$ for \emph{Office Loop} and \emph{Neighborhood}, $1800$ for \emph{Cambridge} and \emph{7Scenes}, except for $300$ for Pose-Transformer in \emph{7Scenes}. For coordinate systems we follow the standard setup, i.e., we use the default coordinate systems in \emph{7Scenes} and \emph{Cambridge}, and transform the coordinate system of \emph{Office Loop} and \emph{Neighbourhood} from SLAM world to ECEF (Earth-centered, Earth-fixed) using 4Seasons official tools~\citep{wenzel2020fourseasons}.

\subsection{Total time consumption}\label{sec:total-time}
In the main experiment, the computation budget is calculated based on the convergence time needed for the initial APR training. In this part, we compute the total update time of all rounds based on this full computation budget. Take \emph{Office Loop} and PT in Tab.~\ref{Tab-condition} for example, each scan has similar number of images. The time consumption spent on each scan is roughly the same ($\approx 20h$). Hence, \emph{ConDo} updates on 3 scans took roughly $3 \times 20h = 60h$; and \emph{Re-train with GT} took $(4+5+6) \times 20h = 300h$ for iterative updates. As analyzed in Table.\ref{Tab-total-time}, ConDo has similar results as \emph{Re-train with GT} while being $5$x faster.
\begin{table}[!h]
\centering
\resizebox{\linewidth}{!}{
\begin{tabular}{c|c|cc|cc}
\hline
\multirow{2}{*}{Update time} & \multirow{2}{*}{Strategy} & \multicolumn{2}{c|}{Training scan held-out data}             & \multicolumn{2}{c}{Inference scan held-out data}               \\
                                    &                            &  Median    &  Mean      &  Median     &  Mean        \\ \hline
 0                                  &  Train-only                 &  1.70/0.29 &  1.82/0.84 &  6.12/16.14 &  42.15/43.24 \\
60h                                &  \cellcolor{gray!40}ConDo                     &  \cellcolor{gray!40}1.34/0.21 &  \cellcolor{gray!40}1.45/0.49 &  \cellcolor{gray!40}1.50/0.49  &  \cellcolor{gray!40}1.86/1.31   \\
300h                               &  Re-train         &  1.41/0.19 &  1.39/0.64 &  1.46/0.18  &  1.59/0.63   \\ \hline
\end{tabular}
}
\caption{Total time costs. Position ($m$) / orientation ($^\circ$) errors of Pose-Transformer in \emph{Office Loop}. As reported in Table.1, we train ConDo and Re-train with GT until convergence in each round and accumulate the time consumption of each round.}
\label{Tab-total-time}
\end{table}

\subsection{Multi-scene Design}\label{sec:ms-design}
\noindent Table.~\ref{Tab-multi-scene-design} shows results on different multi-scene architecture designs of ConDo. As mentioned in Sec.~\ref{sec:exp},  ~\citep{shavit2021learning,shavit2023coarse} learns latent scene embeddings through encoder-decoder attention, but it is only designed for Transformer networks and not available for common APRs (e.g. PoseNet). ~\citep{brachmann2019expert} directly adds manually designed scene position bias to the pose in order to physically separate different scenes but gets suboptimal localization results. We simply add extra regression heads to cope with multi-scene coordinates, which achieves the balance between localization efficiency and network compatibility.
\begin{table}[h]
\centering
\resizebox{\linewidth}{!}{
\begin{tabular}{c|c|cc|cc}
\hline
\multirow{3}{*}{Model} & \multirow{3}{*}{Strategy} & \multicolumn{4}{c}{\emph{7Scenes}}                                                                        \\
                       &                           & \multicolumn{2}{c|}{Training scan held-out data} & \multicolumn{2}{c}{Inference scan held-out data} \\
                       &                           & Median                 & Mean                   & Median                  & Mean                   \\
\hline
\multirow{3}{*}{PN}    & Multi-head                & 0.069/2.198            & 0.078/2.460            & 0.080/2.641             & 0.097/3.042            \\
                       & Latent embed              &     --               &   --                 &          --           &          --         \\
                       & Position bias            & 0.192/2.251            & 0.211/2.542   
         & 0.210/2.570             & 0.231/3.024             \\
\hline
\multirow{3}{*}{PT}    & Multi-head                & 0.049/1.581            & 0.054/1.885            & 0.065/2.113             & 0.077/2.381            \\
                       & Latent embed              & 0.043/1.520            & 0.048/1.745            & 0.058/1.990             & 0.076/2.406            \\
                       & Position bias            & 0.067/1.456            & 0.079/1.826            & 0.081/1.931             & 0.096/2.228            \\
\hline
\end{tabular}
}
\caption{Results on different multi-scene architectures. ``--" means not available. Results are reported in the form of position ($m$) / orientation ($^\circ$) errors.}\label{Tab-multi-scene-design}
\end{table}

\subsection{Further analysis on data with pose changes}\label{sec:multi_scene_performance_drop}
\noindent In the main experiments (Table.~\ref{Tab-pose}), we find that adding inference scan data with obvious pose changes may not always help the performance of training scans. Here, we further analyze this phenomenon by constructing more experiments. There are several possible reasons for the performance decay on the training scans.
\begin{enumerate}
    \item The noise introduced in knowledge distillation of ConDo, since we do not leverage any ground-truth on unlabeled data.
    \item Training on more data with strong pose changes interferes the performance on training scans.
    \item Learning to localize multiple scenes using a single APR model might introduce a negative impact due to cross-scene interference.
    \item The sequential learning of ConDo, i.e., instead of training on all scans together, the sequentially revealed data might hurt the convergence of the model.
\end{enumerate}

For the first factor, we compare \emph{Re-train with GT} and \emph{Re-train with HLoc} in Table~\ref{Tab-HLoc-all}, where \emph{Re-train with HLoc} simply replace the supervision signal on inference scans in \emph{Re-train with GT} with the distillation loss used in ConDo. The results show that \emph{using distillation has minimal impact to the performance drop on training scans.} 

For the second factor, we compare \emph{Train-only} and \emph{Re-train with GT} in Table~\ref{Tab-HLoc-all}, which are the models trained respectively on training scans and training scans plus inference scans. We can see that unlike the case of scene condition change (Table.~\ref{Tab-condition} of the main paper), \emph{Re-train with GT} did not improve the performance on training scans, which shows that \emph{training on more data with strong pose changes indeed has a negative impact} to the performance of individual scan, regardless of whether ConDo is applied. 

For the third factor, we train$\slash$update per-scene APR models in Table~\ref{Tab-stand-alone}, where instead of using multiple heads, we use a completely separate model for each scene. Comparing to the results in Table~\ref{Tab-HLoc-all}, we see that \emph{the multi-head architecture, though more scalable in practice, does have negative impact} to the APR model, especially to ConDo.

For the final factor, we compare \emph{ConDo} and \emph{Re-train with HLoc} in Table~\ref{Tab-HLoc-all}. The results show that \emph{sequential learning in ConDo also contributes to the performance drop on training scans, especially when multi-head architecture is used}.

Hence, we conclude that APR models do not always benefit from learning on more data with pose changes or from new scenes. This is due to mainly three factors, 1) data with strong pose changes or from completely new scenes interfere the performance of APR in general, regardless of whether ConDo is applied. 2) Perform APR for multiple scenes using a compact multi-head architecture hurts the performance in general, regardless of whether ConDo is applied. 3) The sequential learning of ConDo. This result shows that designing APR architectures that benefit from seeing more data in general, including data with strong pose changes, is an important future work in practice. Nonetheless, ConDo still significantly improved the performance on inference scans, reaching a reasonable balance on the performance of all scans (training and inference).
\begin{table}[h!]
\centering
\resizebox{\linewidth}{!}{
\begin{tabular}{c|c|cc|cc}
\hline
\multirow{3}{*}{Model} & \multirow{3}{*}{Strategy} & \multicolumn{4}{c}{\emph{7Scenes}}                                                                        \\
                       &                           & \multicolumn{2}{c|}{Training scan held-out data} & \multicolumn{2}{c}{Inference scan held-out data} \\
                       &                           & Median                 & Mean                   & Median                  & Mean                   \\
\hline
\multirow{4}{*}{PN}    & Train-only                & 0.023/0.925            & 0.027/1.236            & 0.303/9.536             & 0.374/15.228           \\
                       & ConDo                     & 0.069/2.198    & 0.078/2.460    & 0.080/2.641     & 0.097/3.042    \\
                       & Re-train with GT          & 0.024/0.995            & 0.029/1.265            & 0.024/1.016             & 0.028/1.191            \\
                       & Re-train with HLoc        & 0.025/0.998            & 0.030/1.280            & 0.042/1.616             & 0.055/1.929            \\
\hline
\multirow{4}{*}{PT}    & Train-only                & 0.021/1.158          & 0.025/1.547          & 0.198/8.373          & 0.284/12.162           \\
                       & ConDo                     & 0.049/1.581          & 0.054/1.885          & 0.065/2.113          & 0.077/2.381            \\
                       & Re-train with GT          &0.026/1.267          & 0.030/1.535        &  0.026/1.261         &  0.030/1.507            \\ 
                       & Re-train with HLoc        & 0.028/1.303            & 0.031/1.594            & 0.040/1.746             & 0.055/2.140           \\
\hline
\end{tabular}
}
\caption{Comparison with \emph{Re-train with HLoc.} Results are reported in the form of position ($m$) / orientation ($^\circ$) errors.}\label{Tab-HLoc-all}
\end{table}
\begin{table}[h!]
\centering
\resizebox{\linewidth}{!}{
\begin{tabular}{c|c|cc|cc}
\hline
\multirow{3}{*}{Model} & \multirow{3}{*}{Strategy} & \multicolumn{4}{c}{\emph{7Scenes}}                                                                        \\
                       &                           & \multicolumn{2}{c|}{Training scan held-out data} & \multicolumn{2}{c}{Inference scan held-out data} \\
                       &                           & Median                 & Mean                   & Median                  & Mean                   \\
\hline
\multirow{3}{*}{PN}    & Train-only                & 0.019/1.210            & 0.022/1.577            & 0.181/8.542             & 0.252/11.058           \\
                       & ConDo                     & 0.029/1.335            & 0.032/1.661            & 0.048/1.903             & 0.062/2.253            \\
                       & Re-train with GT          & 0.023/1.342            & 0.027/1.701            & 0.023/1.477             & 0.026/1.692            \\
\hline
\multirow{3}{*}{PT}    & Train-only                & 0.015/0.877            & 0.020/1.179            & 0.244/10.013            & 0.322/15.618           \\
                       & ConDo                     & 0.026/1.148            & 0.033/1.431            & 0.046/1.742             & 0.061/2.095            \\
                       & Re-train with GT          & 0.024/1.253            & 0.027/1.646            & 0.025/1.242             & 0.027/1.443             \\
\hline
\end{tabular}
}
\caption{Results of APRs trained stand alone. Results are reported in the form of position ($m$) / orientation ($^\circ$) errors.}\label{Tab-stand-alone}
\end{table}

\subsection{Trajectories Visualization}\label{sec:more-traj}
\begin{figure}[!ht]
    \centering
    \includegraphics[width=\linewidth]{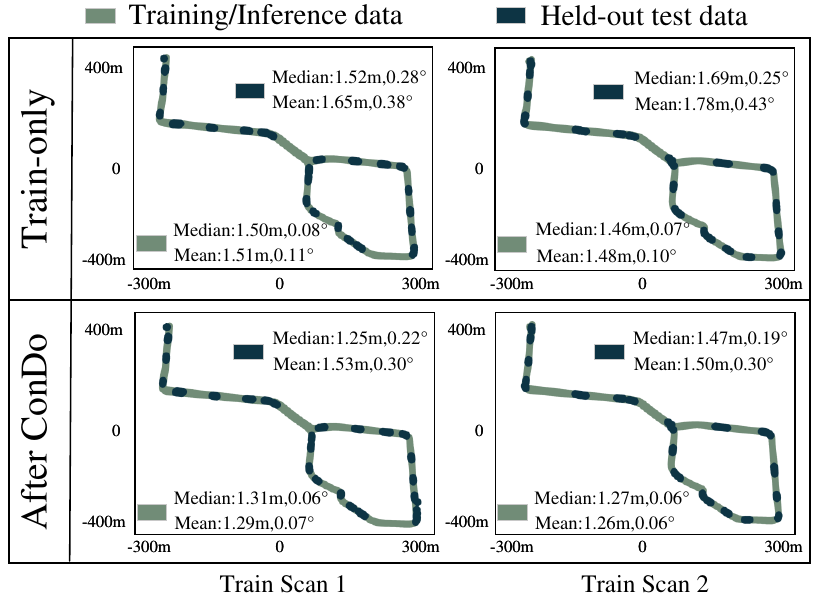}
    \caption{Trajectories visualization on Train Scan 1 and 2.}
    \label{Fig-train-scan}
\end{figure}

In the main paper (Fig.~\ref{Fig-scan-results}), only the results of \emph{Train Scan 3} is visualized due to the space limits. To present more training scan results, we show  \emph{Train Scan 1} and  \emph{2} and their comparison before/after ConDo in Fig.~\ref{Fig-train-scan}. Similar to the case of the main result, ConDo improved held-out data of  \emph{Train Scan 1} from $1.52m$ to $1.25m$ and  \emph{Train Scan 2} from $1.69m$ to $1.47m$, which shows the general robustness improvement of ConDo after updating with unlabelled inference data.

\subsection{Performance of each trajectory at each round}\label{sec:rounds}
\begin{figure}[!ht]
    \centering
    \includegraphics[width=\linewidth]{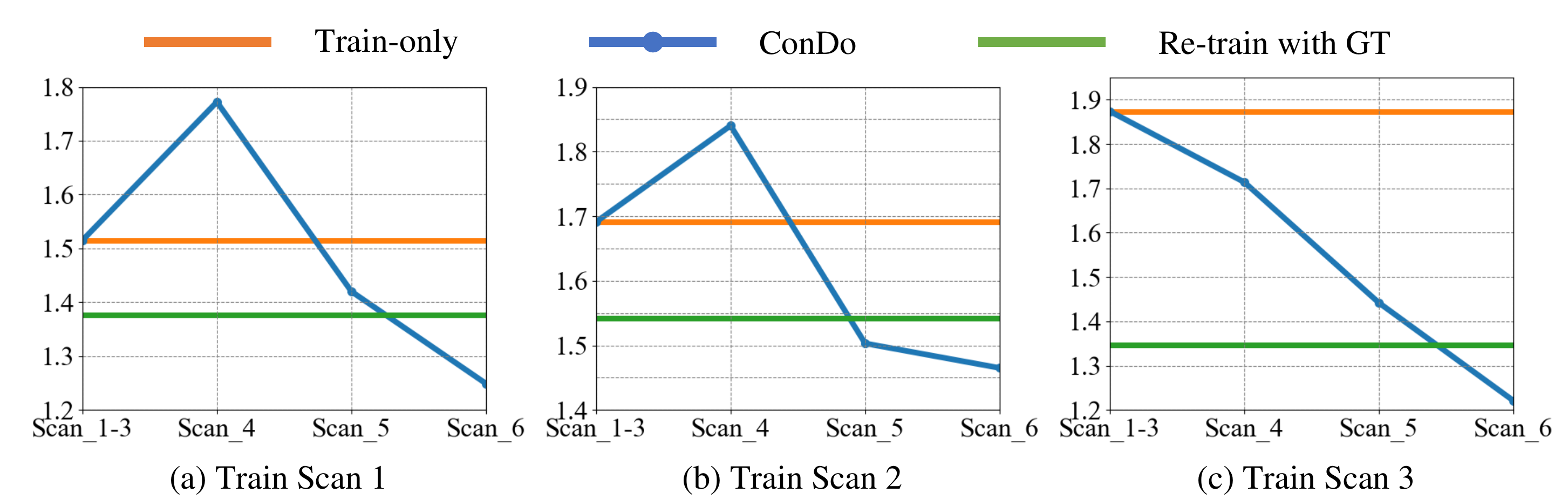}
    \caption{The performance of ConDo on training scans after each round of model update.}
    \label{Fig-train-scan-line}
\end{figure}

In the main paper (Fig.~\ref{Fig-line-chart}), the median position error on held-out data of training scans are illustrated together (Train Scan 1-3) due to the space limit. To better show the performance of ConDo on training scans after each round, we separately report the performance on held-out data of training scans after each round in Fig.~\ref{Fig-train-scan-line}. Results are consistent with the main paper, which indicates that seeing more data during ConDo can improve the general robustness of APR, resulting in a steady accuracy improvement.
\subsection{Teacher settings in Table.\ref{Tab-teacher-effects}}\label{sec:teacher-performance}
For SLAM-type methods (DM-VIO~\cite{von2022dm} and ORB-SLAM~\cite{campos2021orb}), we run their official implementations\cite{von2022dm} on each inference scan to find the relative pose of each frame w.r.t. the first one, and then use the absolute pose of the first frame to get the final supervision signal, i.e., the absolute poses of all other frames. For retrieval-based methods (NetVlad~\cite{arandjelovic2016netvlad}), we use the pose of the retrieved image for supervision.

\subsection{Backbone pre-train.}\label{sec:dino}
Utilizing pre-trained backbones is an effective way to improve generalization of vision models~\citep{keetha2023anyloc,kappeler2023few}. A natural question is whether the generalization problem of APR can be simply addressed by using strong pre-trained backbones. To answer this question, we replace the EfficientNet backbone of PT with a pre-trained Dino v2~\citep{oquab2023dinov2}. Specifically, we use a pre-trained ViT-L/14 to extract patch tokens and GeM (Generalized Mean) Pooling to get 1024-dim features which will be sent to the pose regressor as usual APRs. Then, we train this APR model with training data and decrease the learning rate of Dino v2 to $\frac{1}{10}$ ($10^{-5}$) for better convergence. As shown in Table~\ref{Tab-Dinov2}, introducing Dino v2 (\emph{Train Only} + Dino v2) improved the generalization of APR. However, the performance on challenging data is still low, resulting in a large mean position error ($21m$) indicating the existence of severely failed predictions. ConDo without Dino v2 already achieved a much better performance comparing to \emph{Train Only} + Dino v2. Combining Dino v2 with ConDo further reduced the position error. Hence, naively applying strong pre-trained backbones cannot completely resolve the issue of scene condition and pose changes in APR, though it can complement ConDo and provide performance improvement.
\begin{table}[!ht]
\centering
\resizebox{\linewidth}{!}{
\begin{tabular}{c|c|cc|cc}
\hline
\multirow{2}{*}{Model} & \multirow{2}{*}{Strategy}  & \multicolumn{2}{c|}{Training scan held-out data} & \multicolumn{2}{c}{Inference scan held-out data}  \\
                       &                           & Median               & Mean                 & Median               & Mean    \\
\hline
\multirow{2}{*}{PT}    & Train-only                & 1.70/0.29            & 1.82/0.84           & 6.12/16.14          & 42.15/43.24              \\
                       & ConDo                     & 1.34/0.21            & 1.45/0.49            & 1.50/0.49            & 1.86/1.31             \\
\hline
\multirow{2}{*}{Dinov2}    & Train-only                & 1.73/0.29            & 2.04/0.72           & 4.62/1.27           & 21.38/10.72            \\
                       & ConDo                     & 0.80/0.24            & 1.03/0.37            & 1.02/0.53            & 1.88/1.59             \\
\hline
\end{tabular}
}
\caption{Effect of backbones. Results are evaluated on Office Loop and reported in the form of position ($m$) / orientation ($^\circ$) errors. We replace the EfficientNet backbone of PT with a pre-trained Dino v2 and use the same hyperparameters for the fair comparison, except using $\frac{1}{10}$ learning rate of Dinov2 backbone for convergence. The Dino v2 architecture and pre-training weights are provided by the official released code~\citep{oquab2023dinov2}.}\label{Tab-Dinov2}
\end{table}

\end{document}